\RequirePackage[T1]{fontenc}
\documentclass[letterpaper,journal]{IEEEtran}
\usepackage{amsmath,amssymb}
\usepackage{algorithm,algorithmic}

\usepackage{array,graphicx}
\usepackage{booktabs,tabularx,threeparttable,pifont}
\usepackage{textcomp,stfloats}
\usepackage{cite}
\usepackage{ifthen,etoolbox,xcolor}
\usepackage{url}
\usepackage[hidelinks]{hyperref}
\usepackage[nameinlink,capitalize]{cleveref}
\hypersetup{pdfauthor={},
  pdftitle={DiffDreamer: Coupling World Models with Differentiable Simulation for Agile Quadrotor Control}}
\crefname{figure}{Fig.}{Figs.}
\Crefname{figure}{Fig.}{Figs.}
\crefname{section}{Section}{Sections}

\newboolean{debug}
\setboolean{debug}{false}
\newcommand{\msout}[1]{}

\title{DiffDreamer: Coupling World Models with Differentiable Simulation for Agile Quadrotor Control}

\newif\ifAnonymousReview
\author{\IEEEauthorblockN{Fanxing Li, Yuxiang Huang, Shengyang Wang, Fangyu Sun, Shuyu Wu, Tianbao Zhang, Dexin Zuo, Yan Yufei, Danping Zou, Wenxian Yu}}

\begin{document}
\maketitle

\begin{abstract}

Tasks such as quadrotor dynamic-gate racing require agents to anticipate external motion while maintaining precise high-frequency control. World-model-based reinforcement learning (RL) provides a natural solution by learning latent dynamics for future imagination, yet it often entangles environmental evolution and the robot's own dynamics in a single learned rollout, where model bias can affect policy optimization efficiency. We propose \textbf{DiffDreamer}, a differentiable reinforcement learning framework that tightly couples a learned world model with differentiable simulation for actor optimization. In one optimization step, the learned world model propagates latent states that capture external dynamics and history-dependent features, while the differentiable simulator propagates the robot’s explicit physical state.  We further decompose the actor objective into a critic-driven pathway for non-differentiable task rewards and a simulator-driven pathway for differentiable motion rewards. Experiments on aggressive tracking and dynamic-gate racing show that DiffDreamer improves training convergence over the model-free, model-based, and differentiable-RL baselines, and final performance over these baselines and classical planning and control methods. Additional experiments assess reward decomposition and the odometry-conditioned relative-state representation through ablations. We also provide proof-of-concept real-world flight experiments for validation.

\end{abstract}



\section{Introduction}

Many robotic tasks require agents to anticipate external environmental evolution to achieve precise control. This structure is particularly prominent in agile aerial robotics. A representative example is dynamic-gate racing with quadrotors \cite{kaufmann2018deep, song2022policy, romer2025flying}, where the robot must predict the future motion of moving gates and generate accurate control commands to pass through narrow openings at high speed. Existing solutions \cite{song2022policy} often address this challenge through modular pipelines that decouple prediction, planning, and control. Although such designs are interpretable and effective, they usually rely on task-specific engineering and optimize each module separately.

World-model-based reinforcement learning \cite{hafner2019dream, hafner2020mastering} provides an integrated alternative. By learning compact latent dynamics from interaction, methods such as Dreamer \cite{hafner2019dream} can imagine future trajectories and optimize policies in latent space. However, in agile robotic control, a learned world model often has to capture both the evolution of the external environment and the robot's high-frequency rigid-body dynamics within the same learned rollout. This entangles two sources of dynamics with different structures: the external dynamics are partially observable and history-dependent, whereas the robot's internal dynamics are often smooth and available through physical modeling. As a result, policy optimization through purely learned rollouts may suffer from model bias and less accurate gradients for precise physical control. This motivates using differentiable simulation for the robot's physical motion, so that the learned world model can focus on latent external dynamics.

Differentiable reinforcement learning (DiffRL) \cite{xu_accelerated_2022, mora2021pods} offers a complementary direction. By backpropagating through a differentiable physical model, it can provide low-variance first-order gradients for policy optimization and be particularly effective for smooth robotic systems. However, differentiable RL alone typically assumes that the transition and reward structures are differentiable and explicitly available. This makes it difficult to handle partially observable external dynamics, moving targets with unknown future motion, and non-differentiable task events such as gate-crossing rewards.

\begin{figure*}[!t]
    \centering
    \includegraphics[width=\linewidth]{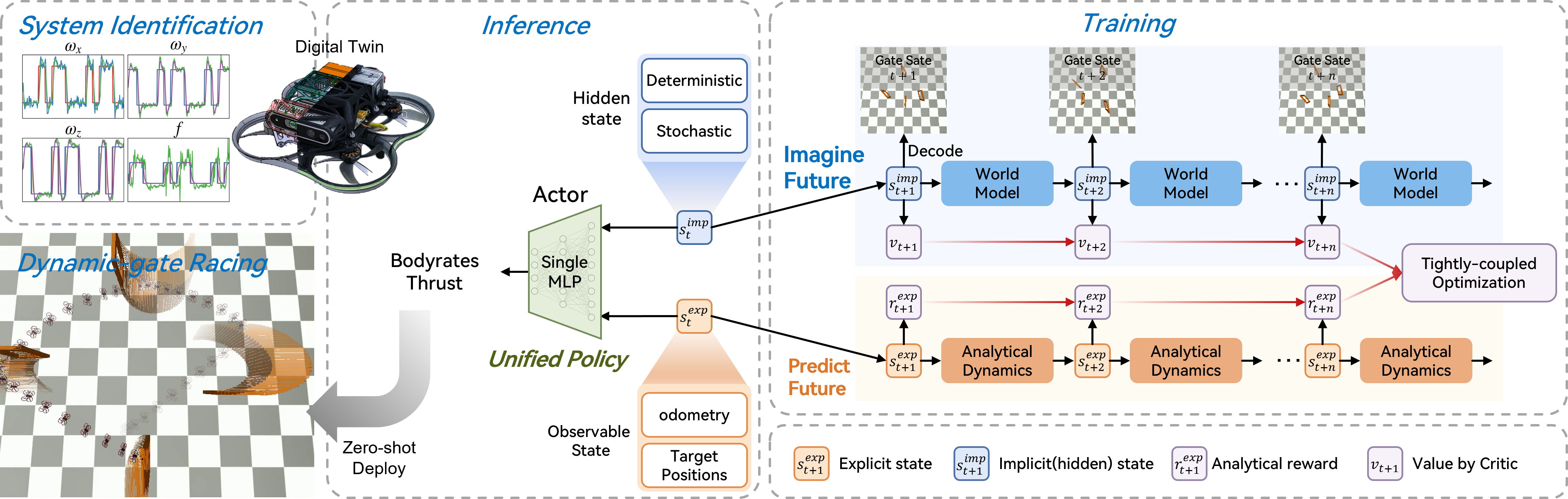}
    \caption{A schematic overview of DiffDreamer for dynamic-gate racing. The quadrotor must predict the future motion of dynamic gates while executing precise flight control. DiffDreamer trains the actor through a tightly coupled dual-pathway process: a learned world model imagines future environmental states (e.g., target motion), while a differentiable simulation predicts the quadrotor's physical dynamics and provides analytical gradients. The policy takes both implicit latent states and explicit physical states as input to generate control commands. The quadrotor is digitally reconstructed and calibrated through system identification for zero-shot transfer to the real world.}
    \label{fig:overview}
    \vspace{-12pt}
\end{figure*}

To address this problem, we propose DiffDreamer, which structurally separates robot dynamics from external environment dynamics and couples their rollouts for policy optimization. DiffDreamer synchronizes latent world-model imagination with differentiable physical rollouts: under the same action, the world model updates an implicit state representing external time-varying features, while the simulator propagates the explicit robot state. The actor receives both states as input, and its objective is decomposed into a critic-driven pathway for non-differentiable task rewards and a simulator-driven pathway for differentiable motion rewards. We compare DiffDreamer with RL, trajectory planning, and model predictive control (MPC) baselines on aggressive tracking and dynamic-gate racing, and demonstrate its effectiveness through transfer to real quadrotor flight.

Our contributions are summarized as follows:
\begin{itemize}
    \setlength{\itemsep}{0pt}
    \setlength{\parsep}{0pt}
    \item We propose a dual-pathway policy optimization framework that synchronizes latent imagination with differentiable physical rollouts to improve optimization accuracy and reduce the burden of learning high-frequency robot dynamics within the world model.
    \item We conduct comprehensive experiments on two quadrotor tasks with dynamic external environments to validate the advantages of the proposed method in terms of control accuracy and predictive capabilities.
    \item We conduct ablation studies on reward decomposition and odometry-conditioned relative-state representation to assess their contributions.
    \item We provide proof-of-concept validation of aggressive tracking and dynamic-gate racing in real-world scenarios.
\end{itemize}

\section{Related Work}

\subsection{Agile Quadrotor Flight and Modular Pipelines}

Agile quadrotor flight, including autonomous racing, typically relies on modular pipelines for perception, state estimation, trajectory planning, and low-level control under tight real-time constraints~\cite{hanover2024autonomous}. In dynamic environments with moving targets or gates, these pipelines often use motion-prediction modules to support downstream planning and control.
Representative approaches combine learned gate perception with trajectory generation and tracking~\cite{kaufmann2018deep}, learned data abstraction with nonlinear filtering and time-optimal planning~\cite{foehn2022alphapilot}, or model-predictive and time-optimal control for aggressive waypoint or gate traversal~\cite{foehn2021time, romero2022model, krinner2024mpcc}.
Policy search can also tune model-predictive controllers for agile flight~\cite{song2022policy}.
Elastic Tracker combines target-motion prediction with visibility-aware spatio-temporal trajectory optimization for aerial tracking~\cite{ji2022elastic}.
These modular pipelines are effective and interpretable, especially when the environment structure and task objectives are well specified.
However, their modules are usually designed and optimized separately, so prediction or planning errors may propagate to downstream control.

\subsection{Model-based Reinforcement Learning}

Model-based reinforcement learning uses learned or known dynamics models for planning and policy optimization, supporting sample-efficient end-to-end policy learning. Model-free methods learn directly from interaction with the environment and typically require more data~\cite{schulman2017proximal, haarnoja2018soft}. Classical and modern approaches have explored probabilistic dynamics and uncertainty-aware planning, such as PILCO and PETS \cite{deisenroth2011pilco,chua2018deep}, short model-generated rollouts for value/policy learning, such as MBPO \cite{janner2019trust}, and latent dynamics for planning or imagination-based actor-critic learning, such as MuZero, Dreamer, and TD-MPC \cite{schrittwieser2020mastering,hafner2019dream,hafner2020mastering,hafner2023mastering,hansen2022temporal,hansen2023td}.

Despite their sample efficiency, model-based methods remain sensitive to model fidelity, especially in high-accuracy and high-frequency robotic control \cite{m2023model}. In agile quadrotor tasks, a single learned rollout must often capture both partially observable external motion and fast rigid-body dynamics, which introduces distinct learning challenges.

\section{Methodology}
\label{sec:methodology}
\subsection{Preliminaries}

We formulate robotic control as a partially observable Markov decision process (POMDP), \(\mathcal{M}=(\mathcal{S},\mathcal{A},p,r,c)\). In state \(s_t\in\mathcal{S}\), the agent executes an action \(a_t\in\mathcal{A}\), receives a reward $r_t\sim r(s_t,a_t)$ and an episode continuation signal $c_t$, and transitions to $s_{t+1}\sim p(s_{t+1}\mid s_t,a_t)$. Here, $p$ and $r$ denote the transition dynamics and reward function. Actions are sampled from a stochastic policy parameterized by $\theta$:
\begin{equation}
    a_t\sim \pi_\theta(\cdot \mid o_t)
\end{equation}
where $o_t$ is a noisy or partial measurement of the true state $s_t$. The actor-critic framework is widely used in reinforcement learning (RL) for such problems. The critic \(V_\psi(o_t)\) (or \(Q_\psi(o_t,a_t)\)), parameterized by \(\psi\), estimates the expected long-term return and can provide a low-variance gradient signal for actor optimization. The actor aims to infer optimal actions from partial observations.

\subsection{Model-based RL from a World Model}
Unlike model-free RL, model-based RL explicitly learns environmental dynamics and imagines future trajectories to estimate long-term returns, which can provide a richer policy optimization signal. As a representative framework, DreamerV3 \cite{hafner2023mastering} uses a recurrent state-space model (RSSM) as its world model to model environment dynamics.
The RSSM represents environmental states and transitions in a compact latent space comprising a deterministic hidden state \(h_t\) and a stochastic latent feature \(z_t\). An encoder $q_\phi$ maps the high-dimensional observation \(o_t\) to a posterior stochastic representation ${z}_t$. The recurrent dynamics module advances from the previous state using the executed action \(a_t\) to predict the temporal prior $(\hat{z}_t, h_t)$. The model uses the concatenated latent state to predict expected rewards and episode continuation flags and reconstruct the original observations.

\textbf{Training the Actor-Critic.}
DreamerV3 trains its actor and critic entirely on abstract trajectories imagined by the learned world model. Starting from initialized latent states $s_t^{imp}=\{z_t, h_t\}$, it unrolls the recurrent dynamics using successive actions from the current actor policy to generate future latent states and predicted rewards. Both networks operate directly on the implicit latent state $s_t^{imp}$ rather than explicit states:
\begin{equation}
    \begin{aligned}
        \text{Actor: }& a_t \sim \pi_\theta (\cdot \mid s_t^{imp}) \\
        \text{Critic: }& \hat{\mathcal{V}}_t \sim v_\psi(\cdot \mid s_t^{imp})
    \end{aligned}
\end{equation}
The actor is trained to maximize expected cumulative returns along imagined trajectories. We express its objective as loss minimization, using a simplified gradient form that emphasizes gradient flow through imagined latent trajectories:
\begin{equation}
    \nabla_\theta \mathcal{L}(\theta) = - \sum_{t=0}^{T} \gamma^t \nabla_\theta \hat{\mathcal{V}}_t
\end{equation}
where $\gamma$ is the discount factor balancing immediate and future rewards, and $T$ is the imagination horizon. 
The critic regresses its value predictions toward expected returns computed from imagined rewards. To robustly estimate the target return $\hat{\mathcal{R}}_t^\lambda$ while mitigating truncation bias beyond the imagination horizon, bootstrapped $\lambda$-returns smoothly interpolate between $N$-step predicted rewards and learned value estimates~\cite{sutton1998reinforcement}. The critic evaluates the return distribution by minimizing the negative log-likelihood:
\begin{equation}
    \label{eq:wm_train_critic}
    \begin{aligned}
        \mathcal{L}(\psi) &= -\sum_{t=0}^{T} \ln p_\psi(\hat{\mathcal{R}}_t^\lambda \mid s_t^{imp}), \\
        \hat{\mathcal{R}}_t^\lambda &= \hat{r}_t + \gamma c_t\big[(1-\lambda)\hat{\mathcal{V}}_{t+1}
          + \lambda \hat{\mathcal{R}}_{t+1}^\lambda \big]
    \end{aligned}
\end{equation}
Here $c_t$ denotes episode continuation, and $\lambda$ controls the mixture of bootstrapped values and longer-horizon returns.

\subsection{Differentiable Reinforcement Learning}
Although traditional RL has achieved remarkable success across diverse fields, actor-critic performance can be sensitive to errors in approximating long-term returns. In discrete domains such as Go or Atari games, parameterized critics are commonly used for value estimation. Smooth robotic control can exploit differentiable models of Newtonian dynamics together with suitably defined differentiable rewards.

Like MPC, differentiable RL directly optimizes the policy by backpropagating reward gradients through a differentiable physical model. This reduces reliance on a learned critic for dense motion optimization and provides low-variance first-order gradients when dynamics and reward terms are smooth and accurately modeled. We define the robot's explicit state as \(s_t^{\mathrm{exp}}\) (e.g., position, velocity, joint angles) and assume a deterministic transition $d(s_t^{\mathrm{exp}}, a_t)$ rather than a stochastic process. The reward $R(s_t^{\mathrm{exp}}, a_t)$ is likewise a deterministic function of the explicit state and action.

In differentiable simulation, the agent collects trajectories by unrolling the dynamics with its current stochastic policy. The resulting complete computational graph links each timestep's reward to the actor parameters across time, allowing the cumulative return to be differentiated with respect to these parameters by backpropagating through the simulator:
\begin{equation}
    \label{eq:diff}
    \nabla_\theta \mathcal{J}(\theta) = \sum_{t=0}^{T} \gamma^t \nabla_\theta R(s_t^{\mathrm{exp}}, a_t)
\end{equation} 

Since each timestep's reward is continuously differentiable with respect to all preceding actions, the policy can be optimized by applying the chain rule through the dynamics. The actor gradient in \cref{eq:diff} thus expands as:
\begin{equation}
\begin{aligned}
\nabla_{\theta}\mathcal{J}&(\theta)
=
\sum_{t=0}^{T}\gamma^{t}
\Biggl[
\frac{\partial R\left(s_t^{\mathrm{exp}},a_t\right)}{\partial a_t}
\frac{\partial a_t}{\partial\theta} 
\\
&\quad +
\frac{\partial R\left(s_t^{\mathrm{exp}},a_t\right)}{\partial s_t^{\mathrm{exp}}}
\sum_{i=1}^{t}
\left(
\prod_{j=i+1}^{t}
\frac{\partial s_j^{\mathrm{exp}}}{\partial s_{j-1}^{\mathrm{exp}}}
\right)
\frac{\partial s_i^{\mathrm{exp}}}{\partial a_{i-1}}
\frac{\partial a_{i-1}}{\partial\theta}
\Biggr]
\end{aligned}
\end{equation}
The reward derivatives, ${\partial R}/{\partial a_t}$ and ${\partial R}/{\partial s_t^{\mathrm{exp}}}$, follow directly from the known analytical reward formulation. The differentiable physics engine provides the state transition Jacobian, ${\partial s_j^{\mathrm{exp}}}/{\partial s_{j-1}^{\mathrm{exp}}}$. For the continuous-time dynamics $d(s_t^{\mathrm{exp}}, a_t)$, the simulator typically uses a time-discretized ordinary differential equation (ODE) integration scheme:
\begin{equation}
    s_{t+1}^{\mathrm{exp}} = s_t^{\mathrm{exp}} + d(s_t^{\mathrm{exp}}, a_t) \Delta t
\end{equation}
Using these first-order gradients through the differentiable physical model, DiffRL can improve sample efficiency in smooth robotic control problems, as empirically validated in complex domains including quadrupedal locomotion \cite{schwarke2024learning} and vision-based navigation for aerial robots \cite{zhang2025learning}.

For the quadrotor, the continuous-time dynamics $d$ govern the world-frame position $\mathbf{p}_W$ and velocity $\mathbf{v}_W$, attitude quaternion $\mathbf{q}$, and body angular velocity $\mathbf{\Omega}$:
\begin{equation}
\label{eq:dynamics}
\begin{aligned}
    \dot{\mathbf{p}}_W &= \mathbf{v}_W, \\
    \dot{\mathbf{v}}_W &= \frac{1}{m}\left(\mathbf{R}_{WB}\mathbf{f}_B+\mathbf{d}_W\right)+\mathbf{g}, \\
    \dot{\mathbf{q}} &= \frac{1}{2}\mathbf{q}\otimes \mathbf{\Omega}, \\
    \dot{\mathbf{\Omega}} &= \mathbf{J}^{-1}
    \left(\boldsymbol{\eta}-\mathbf{\Omega}\times\mathbf{J}\mathbf{\Omega}\right).
\end{aligned}
\end{equation}
Here, $\mathbf{R}_{WB}$ maps body-frame vectors to the world frame, $m$ is the vehicle mass, $\mathbf{J}$ the inertia matrix, $\mathbf{g}$ gravity, $\mathbf{d}_W$ the external disturbance force, and $\boldsymbol{\eta}$ the body torque. The operator $\otimes$ denotes quaternion multiplication. The quadrotor's body-frame thrust force is $\mathbf{f}_B=[0,0,f]^T$, where $f$ is the collective thrust.

\subsection{DiffDreamer: Learning Behaviors from Imagination and Prediction}
\label{sec:diffdreamer}
\begin{figure*}[!t]
    \centering
    \includegraphics[width=\textwidth]{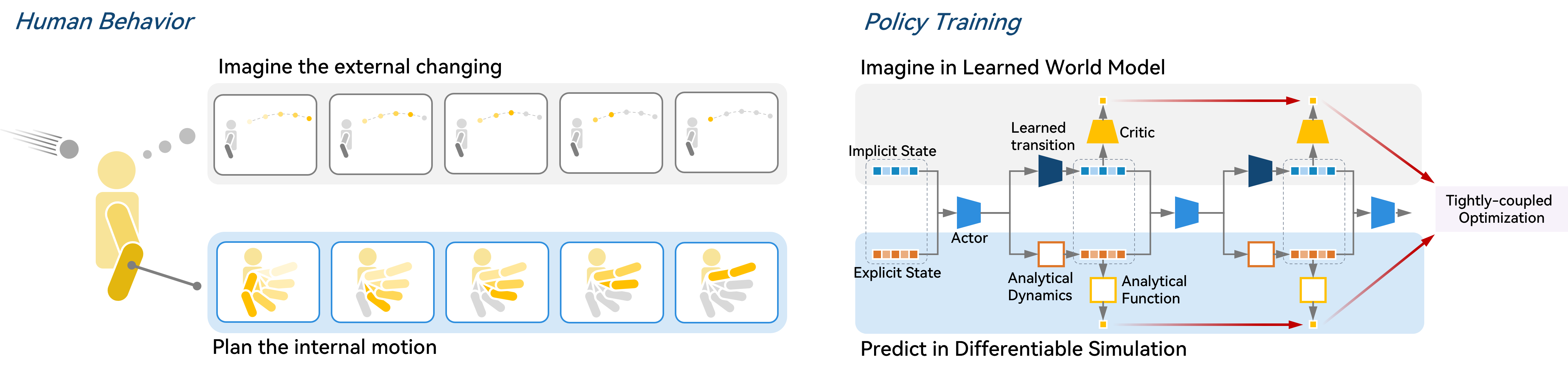}
    \caption{Coupling external-motion prediction with internal control. \textbf{Left:} Catching a ball requires imagining its trajectory and planning body motion. \textbf{Right:} DiffDreamer combines latent world-model features with explicit physical states for policy optimization. It imagines future external motion through a learned world model and propagates the robot's physical state through differentiable simulation. }
    \label{fig:human}
    \vspace{-12pt}
\end{figure*}

Differentiable RL has shown strong sample efficiency in smooth robotic control but requires differentiable dynamics and rewards within the optimization graph. In many dynamic robotic tasks, a calibrated differentiable simulator can model or approximate ego dynamics, while separate prediction or estimation modules typically handle external dynamics such as moving targets or gates. These external models are feasible and often effective, but are usually optimized independently of the control policy, which can contribute to prediction-control mismatch and error propagation at high speeds.

This separation is consistent with the intuitive process of predicting how the external world may evolve while planning body motion using an internal physical model. For example, humans catching an incoming ball anticipate its trajectory while coordinating their body motion, as illustrated in \cref{fig:human}.

Based on this separation, we introduce \textbf{DiffDreamer}, a unified framework coupling predictive latent imagination with differentiable physical prediction. The implicit pathway uses a learned world model to encode history-dependent external dynamics and predict future latent states. The explicit pathway propagates the robot's physical state under the same action through a differentiable simulator, providing gradients for differentiable motion rewards. Actor training through these synchronized pathways combines the latent imagination of world-model-based RL with DiffRL's physical precision.

\textbf{Joint State Actor and Critic}. Like standard model-based RL, DiffDreamer stores interaction experiences in a replay buffer to continually train the world model. The actor and critic use a joint state $s_t$ formed by concatenating the implicit latent state $s_t^{imp}$ and explicit physical state $s_t^{exp}$:
\begin{equation}
    \begin{aligned}
        \text{Actor: }& a_t \sim \pi_\theta (\cdot \mid \mathrm{sg}(s_t^{imp}), s_t^{exp}), \\
        \text{Critic: }& \hat{\mathcal{V}}_t \sim v_\psi(\cdot \mid s_t), \\
        \text{where }& s_t = \{s_t^{imp}, s_t^{exp}\}
    \end{aligned}
\end{equation}
The RSSM produces and updates the implicit state \(s_t^{imp}=(h_t,z_t)\), which encodes history-dependent external dynamics and time-varying features, such as target or gate motion patterns. The differentiable simulator propagates the explicit state \(s_t^{exp}\), comprising position, velocity, attitude, and angular velocity. The stop-gradient operator $\mathrm{sg}$ preserves the actor's latent input features while blocking gradients through that input. At each rollout step, the same actor-sampled action updates both states through the RSSM and simulator, keeping them temporally synchronized.

\textbf{Decomposed Value Estimation}. 
DiffDreamer decomposes rewards into a discrete (or non-differentiable) component $r_t^{disc}=R_{disc}(s_t^{exp},a_t)$ and a differentiable component $r_t^{diff}=R_{diff}(s_t^{exp},a_t)$. Differentiable rewards tightly regulate continuous physical motion, while discrete rewards govern high-level behaviors or task milestones. Rather than approximate the entire expected return, the critic estimates returns from the non-differentiable component; differentiable rewards are optimized by direct backpropagation through the simulator. The critic's surrogate objective in \cref{eq:wm_train_critic} becomes:
\begin{equation}
    \label{eq:diffdreamer_critic}
    \begin{aligned}
        \mathcal{L}(\psi) &= -\sum_{t=0}^{T} \ln p_\psi(\hat{\mathcal{R}}_t^\lambda \mid s_t), \\
        \hat{\mathcal{R}}_t^\lambda &= r_t^{disc} + \gamma c_t\big[(1-\lambda)\hat{\mathcal{V}}_{t+1}
          + \lambda \hat{\mathcal{R}}_{t+1}^\lambda \big]
    \end{aligned}
\end{equation}
Unlike the return in \cref{eq:wm_train_critic}, this return uses only $r_t^{disc}$.

\textbf{Dual-Pathway Policy Optimization}.
The actor's surrogate loss similarly combines a critic-driven value objective with a simulator-driven analytical objective:
\begin{equation}
    \label{eq:diffdreamer_actor}
    \begin{aligned}
        \mathcal{L}(\theta) &= \lambda_{imp} \mathcal{L}_{imp}(\theta)
          + \lambda_{exp} \mathcal{L}_{exp}(\theta), \\
        \mathcal{L}_{imp}(\theta) &= -\sum_{t=0}^{T} \gamma^t \hat{\mathcal{V}}_t, \\
        \mathcal{L}_{exp}(\theta) &= -\sum_{t=0}^{T} \gamma^t r_t^{diff}
    \end{aligned}
\end{equation}
During actor optimization, the world-model and critic parameters are kept fixed. The latent input is detached at each actor evaluation, blocking backpropagation into the imagined latent rollout through this input. The explicit-state input remains differentiable, allowing motion-reward gradients to flow through the physical rollout and the actor. The critic supplies the value-based term for non-differentiable task rewards. Both forward rollouts remain synchronized by the shared action sequence.

DiffDreamer can benefit both paradigms. For model-based RL, separating rigid-body dynamics from latent imagination can reduce the world model's learning burden. For differentiable RL, augmenting physical rollouts with learned latent states extends first-order policy optimization to partially observable dynamic environments and non-differentiable task events. The world model provides latent predictions of external dynamics, while the differentiable simulator supplies gradients for controllable robot dynamics.

\begin{figure*}[!t]
    \centering
    \includegraphics[width=\linewidth]{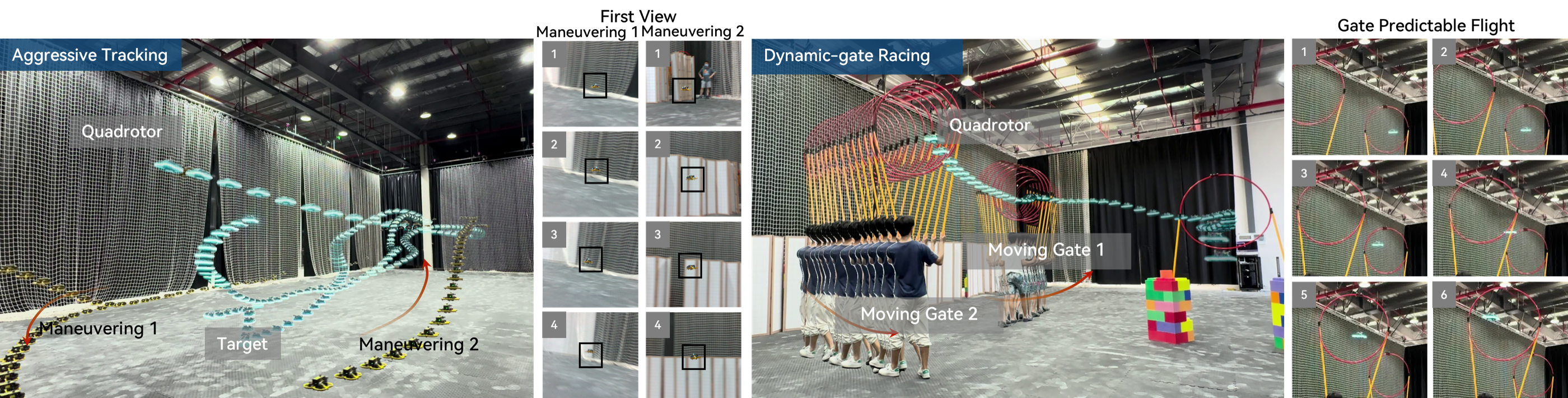}
    \caption{Real-world experiments using DiffDreamer for aggressive tracking and dynamic-gate racing.
\textbf{Left:} flight trajectories and selected consecutive frames showing tight target tracking.
\textbf{Right:} dynamic-gate racing, where the quadrotor anticipates the future positions of moving gates.}
    \label{fig:real_world}
\end{figure*}

\section{Experiments}
Quadrotors' smooth, complex dynamics make them an ideal real-world platform for evaluating algorithms that integrate differentiable physics. We design aggressive tracking and dynamic-gate racing as end-to-end benchmarks to evaluate DiffDreamer's perception of dynamic environments and control performance.

\subsection{Experimental Setup}

We implement both benchmarks in VisFly~\cite{li2025visfly}, a high-fidelity quadrotor simulator with explicit differentiable physics written in PyTorch.
\textbf{Aggressive Tracking} requires the quadrotor to continuously track a high-speed target, keeping it centered in the field of view (FOV) at a stable relative distance.
\textbf{Dynamic-gate Racing} requires the quadrotor to traverse moving gates sequentially in the shortest possible time.

The quadrotor's explicit state $s^{exp}=\{\mathbf{p},\mathbf{v},\mathbf{q},\mathbf{\Omega}\}$ comprises position, velocity, orientation, and angular velocity. We use the full nonlinear rigid-body dynamics in \cref{eq:dynamics}, rather than a point-mass model driven by position commands. The policy outputs end-to-end collective thrust and body rate (CTBR) commands $a_t=[f_t^d,\Omega_t^d]$, where $\Omega_t^d=[\omega_x^d,\omega_y^d,\omega_z^d]$, providing the control authority needed for aggressive maneuvers. Detailed task definitions, observations, and reward formulations are given in \cref{sec:task_definitions}.
Target and gate trajectories are cubic curves with randomly sampled control points.
For external-motion prediction, we compare the world model with a three-dimensional constant-acceleration extended Kalman filter (EKF).
\subsection{Task Definition}
\label{sec:task_definitions}
Observations come from the simulator or, in real-world validation, motion capture, as described in \cref{sec:real_world_deployment}. DreamerV3 encodes observations into its RSSM state; DiffDreamer combines learned latent features with explicit physical states.

\subsubsection{Aggressive Tracking}
The tracking task assumes a monocular camera rigidly mounted on the quadrotor and aligned with the body $x$-axis. The objective is to track a flying target moving agilely and unpredictably in 3D while keeping it centered in the camera field of view (FOV) and maintaining a prescribed constant distance. The observation in this camera-based task combines relative target position with quadrotor odometry:
\begin{equation}
     \label{eq:tracking_observation}
     o=\{\mathbf{p}^{tar}_{H}, \mathbf{v}, \mathbf{q}, \mathbf{\Omega}\}
\end{equation}
where $\mathbf{p}^{tar}_{H}$ denotes the relative position of the target in the heading frame $H$.

We use the StableTracker reward \cite{li2025stabletracker} in VisFly. It combines a distance penalty for deviations from the desired relative distance, a perception reward for facing the target and keeping it centered in the FOV, and a control penalty for aggressive and unstable actions:
\begin{equation}
    \label{eq:reward_func}
    \begin{aligned}
    R(\cdot)& =k_0\frac {\mathbf{x}_B \cdot (\mathbf{p}^{tar}_W-\mathbf{p}_W) }
    {\left \| \mathbf{p}^{tar}_W-\mathbf{p}_W \right \|_2 } \\
    &-k_1 \left | \left \| \mathbf{p}^{tar}_W-\mathbf{p}_W \right \| _2-d \right | \\
    &-k_2\left \| \mathbf{v}_W \right \| _2
    -k_3\left \| \mathbf{\Omega}_W \right \| _2
    +k_4
    \end{aligned}
\end{equation}
where $k_0, k_1, k_2, k_3, k_4$ weight the reward terms and $d$ is the desired distance between the quadrotor and target. The subscript $W$ denotes the world frame. The body-frame $x$-axis vector $\mathbf{x}_B$ coincides with the camera viewing direction.

\subsubsection{Dynamic-gate racing}
Drone racing is a standard benchmark for aerial robots, typically requiring gate traversal as quickly as possible. In our setting, gates move in 3D, requiring the quadrotor to perceive environmental changes and adapt its decisions. Since gates must be traversed sequentially, we include only the nearest two gates in the observation:
\begin{equation}
     \label{eq:racing_observation}
     o=\{\mathbf{p}_{W}^{gate1}-\mathbf{p}_W,\mathbf{p}_{W}^{gate2}-\mathbf{p}_W, \mathbf{v}, \mathbf{q}, \mathbf{\Omega}\}
\end{equation}
where $\mathbf{p}^{gate1}_W-\mathbf{p}_W$ and $\mathbf{p}^{gate2}_W-\mathbf{p}_W$ are the relative positions of the nearest two gates in the world frame.

The reward combines a discrete component for successful gate passage with a differentiable component that encourages motion toward the gate, penalizes motion away from it, and penalizes aggressive and unstable control actions:
\begin{equation}
    \label{eq:reward_func_racing}
    \begin{aligned}
          R_{\mathrm{diff}}(\cdot) &= k_0 \left \|\mathbf{v}_{\text{approach}}\right \|_2 -
          k_1 \left \|\mathbf{v}_{\text{away}}\right \|_2 \\
          &\quad - k_2\left \| \mathbf{\Omega}_W \right \| _2 \\
          R_{\mathrm{disc}}(\cdot) &=
          \begin{cases}
              k_3, & \text{if successfully passed,} \\
              0, & \text{otherwise.}
          \end{cases}
    \end{aligned}
\end{equation}
where $\mathbf{v}_{\mathrm{approach}}$ and $\mathbf{v}_{\mathrm{away}}$ are the velocity components toward and away from the gate, respectively. The coefficients $k_0, k_1, k_2$ weight the differentiable reward terms; $k_3$ is a large positive constant encouraging successful gate passage.

\begin{figure*}[!t]
    \centering
    \includegraphics[width=\linewidth]{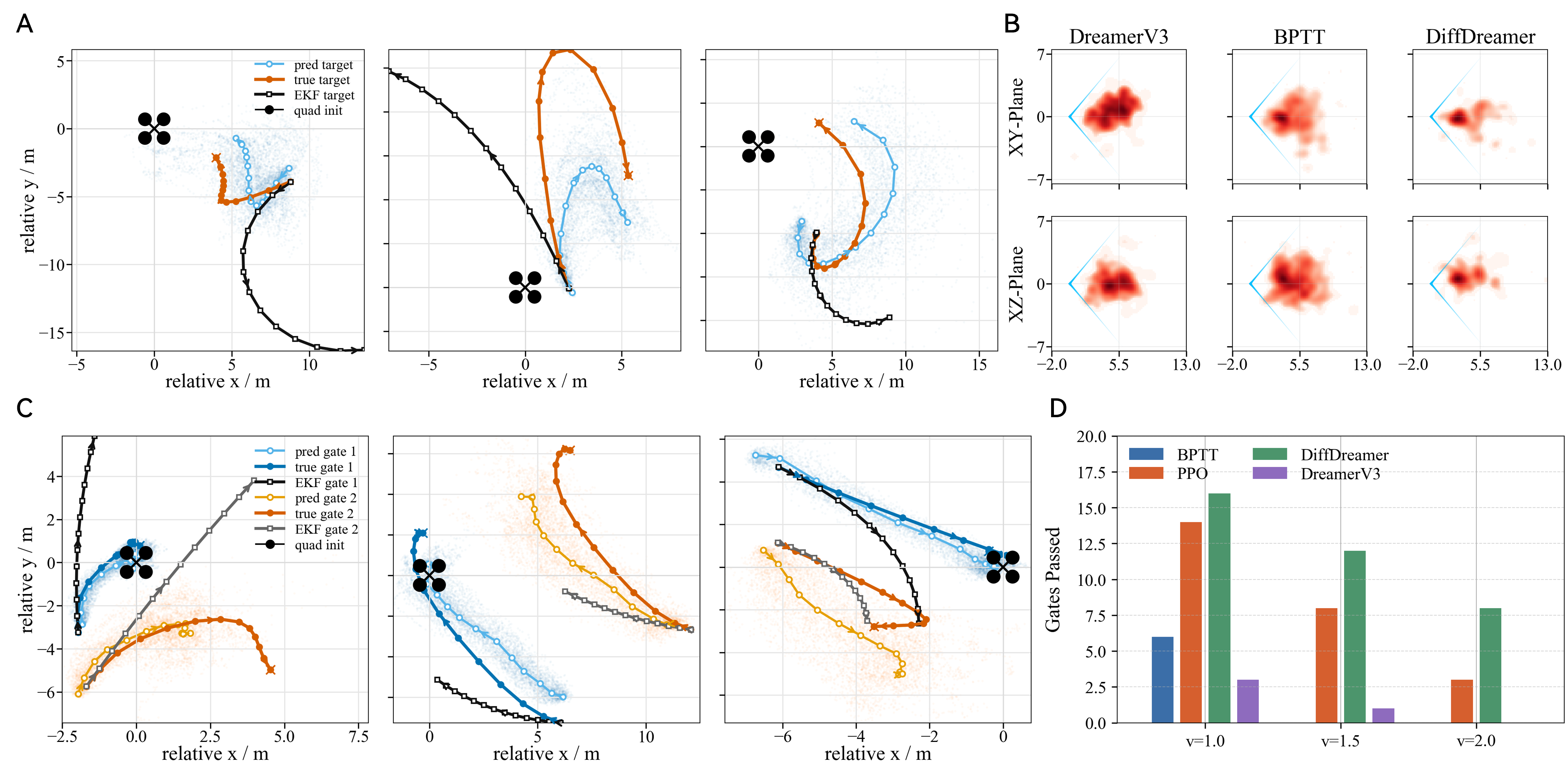}
    
    \caption{The experimental results. 
    \textbf{(A)} The predicted trajectory of the target over a 2-second horizon in aggressive tracking. The scattered points denote imagined trajectory samples. 
    \textbf{(B)} Heatmaps illustrating the spatial distribution of the target within the quadrotor's body frame. More intense colors correspond to higher spatial density. The region bounded by the blue rays represents the camera's actual field of view (FOV).
    \textbf{(C)} The predicted trajectories of the next two gates over a 2-second horizon in dynamic-gate racing.
    \textbf{(D)} The average number of successfully passed gates during dynamic-gate racing.
    }
    \label{fig:sim_exp}
    \vspace{-12pt}
    
\end{figure*}

\subsection{Training Results}
\label{sec:training}
\begin{figure}[!t]
    \centering
    \includegraphics[width=\linewidth]{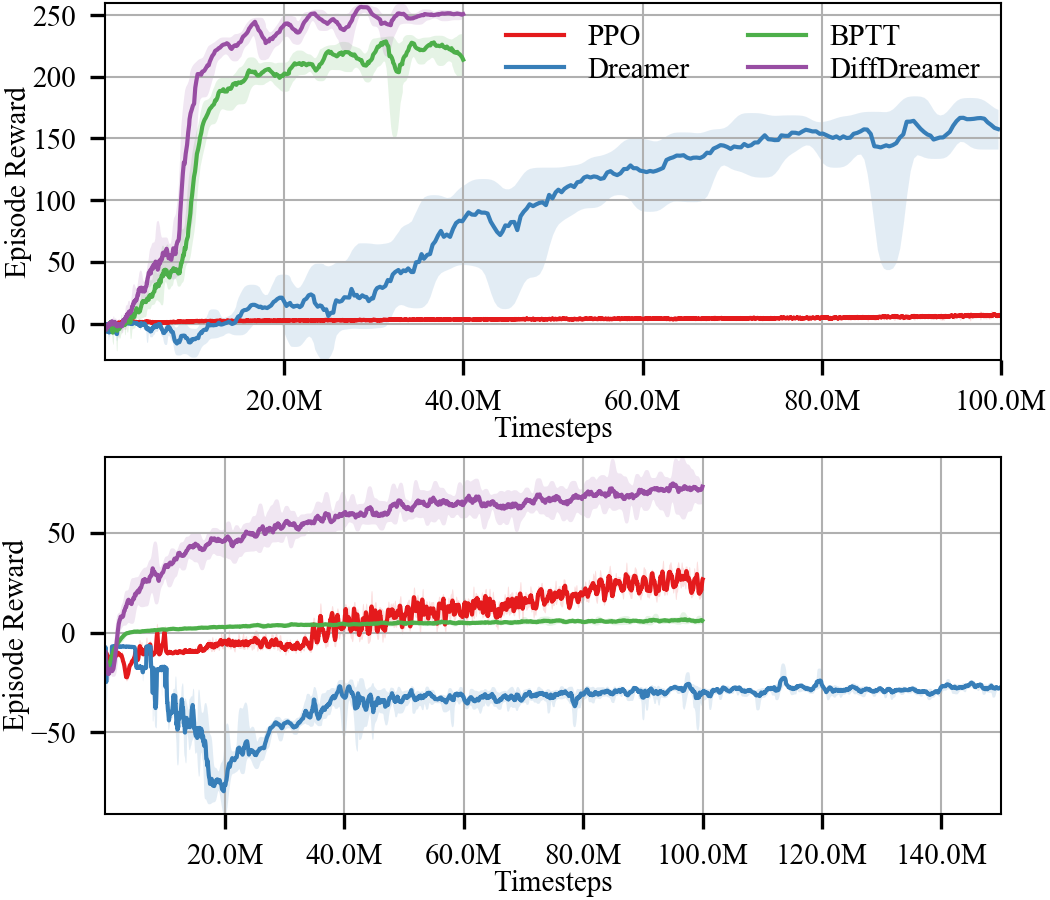}
    \caption{Training reward curves for aggressive tracking (top) and dynamic-gate racing (bottom). Solid lines and shaded regions represent the mean and standard deviation across five independent random seeds, respectively.}
    \label{fig:reward_curve}
    \vspace{-12pt}
\end{figure}

On both benchmarks, we train DiffDreamer, DreamerV3, proximal policy optimization (PPO), and backpropagation through time (BPTT), evaluating each across five random seeds to account for statistical variance. \Cref{fig:reward_curve} shows the training reward curves.
All methods share the same observations, rewards, and uniform randomization of target trajectories, speeds, and initial states, with fixed identified physics parameters. Baselines are tuned individually, while methods in the same methodological family share core hyperparameters, including prediction horizons for differentiable methods. 

The baselines and DiffDreamer share a similar training architecture:
DreamerV3 learns the full latent rollout through its world model from these observations; BPTT uses the differentiable quadrotor simulator and differentiable reward terms without a learned latent-imagination pathway; DiffDreamer combines both pathways, using the world model to predict external dynamics and the differentiable simulator to propagate the quadrotor's physical state. 
DiffDreamer converges faster and achieves higher final rewards than the baselines in both tasks, with larger gains in dynamic-gate racing. This suggests that coupled latent imagination and differentiable rollouts particularly benefit tasks requiring both future prediction and precise optimization of robot motion.

\subsection{Experimental Performance}
We select Elastic Tracker~\cite{ji2022elastic} for tracking and the policy-search method~\cite{song2022policy} for racing as classical planning and control baselines.

\textbf{External-motion Prediction.} The world model and EKF observe the first half of each trajectory and forecast the remainder open loop. \Cref{fig:sim_exp}A and C show that DiffDreamer predicts more precise trajectories than the EKF for agile target motion in both tasks. Without measurement updates, the EKF's fixed-motion model cannot anticipate subsequent acceleration or maneuver changes, so its errors grow rapidly during agile maneuvers.

\textbf{Aggressive Tracking.} We evaluate trained policies in ten randomized scenarios with aggressive target maneuvers along diverse, unpredictable trajectories. \Cref{fig:sim_exp}B shows heatmaps of the target's spatial density in the quadrotor's body frame over entire flights. Concentration around the desired relative position indicates that the policy maintains a stable standoff distance and keeps the target within the camera FOV. DiffDreamer's distribution is more compact than those of the baselines, especially in the XZ plane, suggesting greater robustness to abrupt vertical target motion.

\begin{figure*}[!t]
    \centering
    \includegraphics[width=\linewidth]{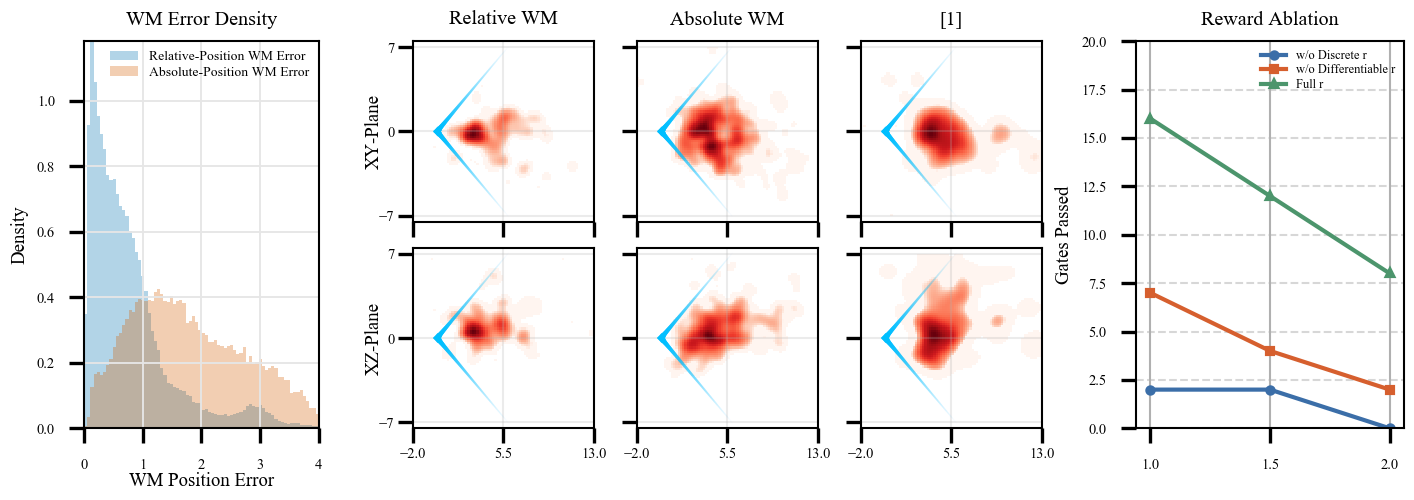}
    \caption{Ablation studies. \textbf{Left and center:} Odometry/representation ablation: reconstruction-error densities and target distributions in the quadrotor's XY/XZ planes. Relative and absolute world models are compared with Elastic Tracker (labeled [1])~\cite{ji2022elastic}; blue rays bound the camera field of view. \textbf{Right:} Reward ablation: gates passed within 512 steps versus maximum gate speed (m/s), using the full reward or removing the discrete gate-passing or differentiable motion component.}
    \label{fig:ablation}
    \vspace{-12pt}
\end{figure*}

DiffDreamer achieves the lowest mean errors on all three tracking metrics in \cref{tab:tracking_mpc}: its mean distance error is $0.381$ m versus $0.793$ m for Elastic Tracker, $0.835$ m for BPTT, and $1.003$ m for DreamerV3. \Cref{fig:ablation} also shows Elastic Tracker's target distribution.

\textbf{Dynamic-gate Racing.}
We evaluate racing in five randomized environments, each with a sequence of four gates moving along varying trajectories. Three difficulty tiers test robustness at maximum gate translation speeds of 1.0, 1.5, and 2.0 m/s. \Cref{fig:sim_exp}D reports the primary metric: the average number of gates passed within 512 simulation steps.

DiffDreamer outperforms all baselines across gate speeds. In \cref{tab:racing_mpc}, its mean gate counts are 16, 12, and 8 at the three speeds, versus 12, 4, and 1 for MPC. This smaller decline relative to MPC suggests greater robustness to faster gate motion. BPTT's performance degrades with gate speed, which may be related to its lack of future motion prediction; DreamerV3's limited performance may reflect the combined complexity of learned physical transitions and sparse gate-passing rewards.

\begin{table}[tb]

\centering
\begin{threeparttable}
\caption{Aggressive tracking errors.}
\label{tab:tracking_mpc}
\footnotesize
\setlength{\tabcolsep}{3pt}
\begin{tabular*}{\columnwidth}{@{\extracolsep{\fill}}lccc@{}}
\toprule
Method & \shortstack{Horizontal\\$|e_h|$} & \shortstack{Vertical\\$|e_v|$} & \shortstack{Distance\\$|e_d|$ (m)} \\
\midrule
DreamerV3 & $0.318\pm0.275$ & $0.314\pm0.307$ & $1.003\pm0.556$ \\
BPTT & $0.198\pm0.227$ & $0.235\pm0.219$ & $0.835\pm0.532$ \\
Elastic Tracker & $0.333\pm0.306$ & $0.382\pm0.305$ & $0.793\pm0.404$ \\
DiffDreamer & $\mathbf{0.168}\pm0.163$ & $\mathbf{0.215}\pm0.177$ & $\mathbf{0.381}\pm0.282$ \\
\bottomrule
\end{tabular*}
\begin{tablenotes}[flushleft]
\footnotesize
\item[] \textit{Notes:} Values are reported as the mean $\pm$ standard deviation over random trajectories and time steps. $|e_h|$ and $|e_v|$ are the normalized horizontal and vertical distances, respectively, from the target's projection to the image center on the image plane; both are dimensionless. $|e_d|$ is the distance error in meters. Lower values are better; bold indicates the best result. Elastic Tracker is the aerial trajectory planner of Ji et al.~\cite{ji2022elastic}.
\end{tablenotes}
\end{threeparttable}

\vspace{-12pt}
\end{table}

\begin{table}[tb]

\centering
\begin{threeparttable}
\caption{Dynamic-gate racing performance.}
\label{tab:racing_mpc}
\footnotesize
\setlength{\tabcolsep}{3pt}
\begin{tabular*}{\columnwidth}{@{\extracolsep{\fill}}lccc@{}}
\toprule
Method & $v=1.0$ & $v=1.5$ & $v=2.0$ \\
\midrule
DreamerV3 & $3.0\pm0.9$ & $1.0\pm0.9$ & $0.0\pm0.0$ \\
BPTT & $6.0\pm3.4$ & $0.0\pm0.5$ & $0.0\pm0.0$ \\
PPO & $14.0\pm2.1$ & $8.0\pm2.3$ & $3.0\pm1.2$ \\
MPC & $12.0\pm0.9$ & $4.0\pm0.9$ & $1.0\pm0.9$ \\
DiffDreamer & $\mathbf{16.0}\pm1.0$ & $\mathbf{12.0}\pm0.5$ & $\mathbf{8.0}\pm1.4$ \\
\bottomrule
\end{tabular*}
\begin{tablenotes}[flushleft]
\footnotesize
\item[] \textit{Notes:} Entries report the number of gates passed within 512 simulation steps as the mean $\pm$ standard deviation over random trajectories and time steps; $v$ is the maximum gate speed in m/s. Higher values are better; bold indicates the best result. MPC uses the approach of Song and Scaramuzza~\cite{song2022policy}.
\end{tablenotes}
\end{threeparttable}

\vspace{-12pt}
\end{table}

\subsection{Ablations}
\label{sec:ablation}

\subsubsection{Odometry and relative-state representation}
\label{sec:odometry_ablation}

The relative-state formulation combines quadrotor odometry with target-relative observations. The ablated world model removes odometry and reconstructs the target's absolute position. In \cref{fig:ablation}, the absolute-position variant has a broader reconstruction-error distribution and a more dispersed target distribution in the quadrotor frame during tracking.

\subsubsection{Reward decomposition}
\label{sec:reward_ablation}

We remove each reward component in turn while evaluating dynamic-gate racing at maximum gate speeds of 1.0, 1.5, and 2.0 m/s. In \cref{fig:ablation}, the full objective yields gate counts of 16, 12, and 8. Removing the discrete gate-passing reward reduces these counts to 2, 2, and 0. Removing the differentiable motion reward yields 7, 4, and 2. Removing either component reduces the number of gates passed at all three speeds.

\subsection{Real-World Experiments}
\label{sec:real_world_deployment}

We evaluate the sim-to-real performance of DiffDreamer policies for both tasks on a custom-built quadrotor with 3.5-inch propellers, an NVIDIA Jetson Orin NX onboard computer, and a Betaflight flight controller. The trained policies run onboard at over 100 Hz, sending collective thrust and body rate (CTBR) commands to the flight controller.
An external motion-capture system provides odometry for the quadrotor and moving targets/gates, allowing us to evaluate DiffDreamer's control and planning with external state feedback.
We carefully identify the quadrotor's physical parameters and incorporate them into the differentiable simulator to reduce the sim-to-real gap. All experiments use bounded thrust commands for safety in a controlled indoor environment. The successful trajectories in \cref{fig:real_world} provide proof-of-concept evidence of transfer to physical flight under calibrated dynamics and motion-capture state feedback.

\section{Conclusion}
We presented DiffDreamer, a differentiable reinforcement learning framework for agile quadrotor tasks that require both external-motion prediction and precise physical control. During actor optimization, DiffDreamer couples two rollouts: the learned world model propagates latent states capturing external dynamics and history-dependent features, while the differentiable simulator propagates the robot's explicit physical state. This separation reduces the burden of learning high-frequency robot dynamics within the world model. The actor objective is decomposed into a critic-driven pathway for non-differentiable task rewards and a simulator-driven pathway for differentiable motion rewards. Experiments on aggressive tracking and dynamic-gate racing show improved training convergence over model-free, model-based, and differentiable-RL baselines, and higher final performance than these baselines and classical planning and control methods. Ablation studies assess the contributions of reward decomposition and the odometry-conditioned relative-state representation. Real-world flight experiments further provide proof-of-concept validation.

\section{Limitations and Discussion}
DiffDreamer has been evaluated only on quadrotors. Extending its coupled latent/physical rollouts and reward decomposition to other embodiments requires suitable differentiable models and additional task-specific engineering. Contacts, discontinuities, actuator limits, and perception modules can complicate differentiation~\cite{metz2021gradients}. In addition, real-world validation relies on motion-capture feedback for both the vehicle and external objects. Fully onboard perception and robustness to perception noise remain untested.

\bibliographystyle{IEEEtran}
\bibliography{reference}

@misc{xu_accelerated_2022,
	title = {Accelerated {Policy} {Learning} with {Parallel} {Differentiable} {Simulation}},
	url = {http://arxiv.org/abs/2204.07137},
	doi = {10.48550/arXiv.2204.07137},
	urldate = {2024-10-12},
	publisher = {arXiv},
	author = {Xu, Jie and Makoviychuk, Viktor and Narang, Yashraj and Ramos, Fabio and Matusik, Wojciech and Garg, Animesh and Macklin, Miles},
	month = apr,
	year = {2022},
	note = {arXiv:2204.07137},
}

@article{chua2018deep,
	title={Deep reinforcement learning in a handful of trials using probabilistic dynamics models},
	author={Chua, Kurtland and Calandra, Roberto and McAllister, Rowan and Levine, Sergey},
	journal={Advances in neural information processing systems},
	volume={31},
	year={2018}
}

@article{janner2019trust,
  title={When to trust your model: Model-based policy optimization},
  author={Janner, Michael and Fu, Justin and Zhang, Marvin and Levine, Sergey},
  journal={Advances in neural information processing systems},
  volume={32},
  year={2019}
}

@article{schrittwieser2020mastering,
  title={Mastering atari, go, chess and shogi by planning with a learned model},
  author={Schrittwieser, Julian and Antonoglou, Ioannis and Hubert, Thomas and Simonyan, Karen and Sifre, Laurent and Schmitt, Simon and Guez, Arthur and Lockhart, Edward and Hassabis, Demis and Graepel, Thore and others},
  journal={Nature},
  volume={588},
  number={7839},
  pages={604--609},
  year={2020},
  publisher={Nature Publishing Group UK London}
}

@article{hafner2019dream,
  title={Dream to control: Learning behaviors by latent imagination},
  author={Hafner, Danijar and Lillicrap, Timothy and Ba, Jimmy and Norouzi, Mohammad},
  journal={arXiv preprint arXiv:1912.01603},
  year={2019}
}

@article{hafner2020mastering,
  title={Mastering atari with discrete world models},
  author={Hafner, Danijar and Lillicrap, Timothy and Norouzi, Mohammad and Ba, Jimmy},
  journal={arXiv preprint arXiv:2010.02193},
  year={2020}
}

@article{hafner2023mastering,
  title={Mastering diverse domains through world models},
  author={Hafner, Danijar and Pasukonis, Jurgis and Ba, Jimmy and Lillicrap, Timothy},
  journal={arXiv preprint arXiv:2301.04104},
  year={2023}
}

@article{hansen2023td,
  title={Td-mpc2: Scalable, robust world models for continuous control},
  author={Hansen, Nicklas and Su, Hao and Wang, Xiaolong},
  journal={arXiv preprint arXiv:2310.16828},
  year={2023}
}

@inproceedings{deisenroth2011pilco,
  title={PILCO: A model-based and data-efficient approach to policy search},
  author={Deisenroth, Marc and Rasmussen, Carl E},
  booktitle={Proceedings of the 28th International Conference on machine learning (ICML-11)},
  pages={465--472},
  year={2011}
}

@article{hansen2022temporal,
  title={Temporal difference learning for model predictive control},
  author={Hansen, Nicklas and Wang, Xiaolong and Su, Hao},
  journal={arXiv preprint arXiv:2203.04955},
  year={2022}
}

@article{m2023model,
  title={Model-based reinforcement learning: A survey},
  author={M. Moerland, Thomas and Broekens, Joost and Plaat, Aske and M. Jonker, Catholijn},
  journal={Foundations and Trends in Machine Learning},
  volume={16},
  number={1},
  pages={1--118},
  year={2023},
  publisher={Emerald Publishing Limited}
}

@article{schulman2017proximal,
  title={Proximal policy optimization algorithms},
  author={Schulman, John and Wolski, Filip and Dhariwal, Prafulla and Radford, Alec and Klimov, Oleg},
  journal={arXiv preprint arXiv:1707.06347},
  year={2017}
}

@inproceedings{haarnoja2018soft,
  title={Soft actor-critic: Off-policy maximum entropy deep reinforcement learning with a stochastic actor},
  author={Haarnoja, Tuomas and Zhou, Aurick and Abbeel, Pieter and Levine, Sergey},
  booktitle={International conference on machine learning},
  pages={1861--1870},
  year={2018},
  organization={Pmlr}
}

@article{schwarke2024learning,
  title={Learning Deployable Locomotion Control via Differentiable Simulation},
  author={Schwarke, Clemens and Klemm, Victor and Bagajo, Joshua and Sleiman, Jean-Pierre and Georgiev, Ignat and Tordesillas, Jesus and Hutter, Marco},
  journal={arXiv preprint arXiv:2404.02887},
  year={2024}
}

@article{zhang2025learning,
  title={Learning vision-based agile flight via differentiable physics},
  author={Zhang, Yuang and Hu, Yu and Song, Yunlong and Zou, Danping and Lin, Weiyao},
  journal={Nature Machine Intelligence},
  volume={7},
  number={6},
  pages={954--966},
  year={2025},
  publisher={Nature Publishing Group UK London}
}

@inproceedings{li2025visfly,
  title={Visfly: An efficient and versatile simulator for training vision-based flight},
  author={Li, Fanxing and Sun, Fangyu and Zhang, Tianbao and Zou, Danping},
  booktitle={2025 IEEE International Conference on Robotics and Automation (ICRA)},
  pages={11325--11332},
  year={2025},
  organization={IEEE}
}

@article{li2025stabletracker,
  title={StableTracker: Learning to Stably Track Target via Differentiable Simulation},
  author={Li, Fanxing and Wang, Shengyang and Sun, Fangyu and Wu, Shuyu and Zuo, Dexin and Yu, Wenxian and Zou, Danping},
  journal={arXiv preprint arXiv:2509.14147},
  year={2025}
}

@inproceedings{kaufmann2018deep,
  title={Deep drone racing: Learning agile flight in dynamic environments},
  author={Kaufmann, Elia and Loquercio, Antonio and Ranftl, Rene and Dosovitskiy, Alexey and Koltun, Vladlen and Scaramuzza, Davide},
  booktitle={Conference on Robot Learning},
  pages={133--145},
  year={2018},
  organization={PMLR}
}

@article{song2022policy,
  title={Policy search for model predictive control with application to agile drone flight},
  author={Song, Yunlong and Scaramuzza, Davide},
  journal={IEEE Transactions on Robotics},
  volume={38},
  number={4},
  pages={2114--2130},
  year={2022},
  publisher={IEEE},
  doi={10.1109/TRO.2022.3141602}
}

@inproceedings{ji2022elastic,
  title={{Elastic Tracker}: A Spatio-temporal Trajectory Planner for Flexible Aerial Tracking},
  author={Ji, Jialin and Pan, Neng and Xu, Chao and Gao, Fei},
  booktitle={2022 International Conference on Robotics and Automation (ICRA)},
  year={2022},
  pages={47--53},
  publisher={IEEE},
  doi={10.1109/ICRA46639.2022.9811688}
}

@inproceedings{romer2025flying,
  title={Flying through Moving Gates without Full State Estimation},
  author={R{\"o}mer, Ralf and Emmert, Tim and Schoellig, Angela P},
  booktitle={2025 IEEE International Conference on Robotics and Automation (ICRA)},
  pages={9906--9912},
  year={2025},
  organization={IEEE}
}

@inproceedings{mora2021pods,
  title={Pods: Policy optimization via differentiable simulation},
  author={Mora, Miguel Angel Zamora and Peychev, Momchil and Ha, Sehoon and Vechev, Martin and Coros, Stelian},
  booktitle={International Conference on Machine Learning},
  pages={7805--7817},
  year={2021},
  organization={PMLR}
}

@article{hanover2024autonomous,
  title={Autonomous drone racing: A survey},
  author={Hanover, Drew and Loquercio, Antonio and Bauersfeld, Leonard and Romero, Angel and Penicka, Robert and Song, Yunlong and Cioffi, Giovanni and Kaufmann, Elia and Scaramuzza, Davide},
  journal={IEEE Transactions on Robotics},
  volume={40},
  pages={3044--3067},
  year={2024},
  publisher={IEEE}
}

@article{foehn2022alphapilot,
  title={Alphapilot: Autonomous drone racing},
  author={Foehn, Philipp and Brescianini, Dario and Kaufmann, Elia and Cieslewski, Titus and Gehrig, Mathias and Muglikar, Manasi and Scaramuzza, Davide},
  journal={Autonomous Robots},
  volume={46},
  number={1},
  pages={307--320},
  year={2022},
  publisher={Springer}
}

@article{foehn2021time,
  title={Time-optimal planning for quadrotor waypoint flight},
  author={Foehn, Philipp and Romero, Angel and Scaramuzza, Davide},
  journal={Science robotics},
  volume={6},
  number={56},
  pages={eabh1221},
  year={2021},
  publisher={American Association for the Advancement of Science}
}

@article{romero2022model,
  title={Model predictive contouring control for time-optimal quadrotor flight},
  author={Romero, Angel and Sun, Sihao and Foehn, Philipp and Scaramuzza, Davide},
  journal={IEEE Transactions on Robotics},
  volume={38},
  number={6},
  pages={3340--3356},
  year={2022},
  publisher={IEEE}
}

@article{krinner2024mpcc,
  title={Mpcc++: Model predictive contouring control for time-optimal flight with safety constraints},
  author={Krinner, Maria and Romero, Angel and Bauersfeld, Leonard and Zeilinger, Melanie and Carron, Andrea and Scaramuzza, Davide},
  journal={arXiv preprint arXiv:2403.17551},
  year={2024}
}

@book{sutton1998reinforcement,
  title={Reinforcement learning: An introduction},
  author={Sutton, Richard S and Barto, Andrew G and others},
  volume={1},
  number={1},
  year={1998},
  publisher={MIT press Cambridge}
}

@article{metz2021gradients,
  title={Gradients are not all you need},
  author={Metz, Luke and Freeman, C Daniel and Schoenholz, Samuel S and Kachman, Tal},
  journal={arXiv preprint arXiv:2111.05803},
  year={2021}
}

\end{document}